\documentclass[12pt]{iopart}

\usepackage{amssymb}  
\usepackage{pifont}

\expandafter\let\csname equation*\endcsname\relax
\expandafter\let\csname endequation*\endcsname\relax

\usepackage{amsmath}
\usepackage{longtable}
\usepackage{booktabs}
\usepackage[a4paper, margin=1in]{geometry}
\usepackage{booktabs}
\usepackage{multirow}
\usepackage{makecell} 
\usepackage{appendix}
\usepackage{graphicx}
\usepackage{url}

\begin{document}

\title[Parallel MARL for automatic HNC TPP tuning]{Large-scale automatic carbon ion treatment planning for head and neck cancers via parallel multi-agent reinforcement learning}



\author{Jueye Zhang$^{1,2,3}$, Chao Yang$^{3}$, Youfang Lai$^{3}$, Kai-Wen Li$^{3}$, Wenting Yan$^{3}$, Yunzhou Xia$^{3}$, Haimei Zhang$^{3}$, Jingjing Zhou$^{3}$,  Gen Yang$^{1}$, Chen Lin$^{1}$, Tian Li$^{4}$ and Yibao Zhang$^{2,*}$}

\address{$^1$ State Key Laboratory of Nuclear Physics and Technology, Institute of Heavy Ion Physics, Peking University School of Physics, Beijing 100871, China.}
\address{$^2$ Key Laboratory of Carcinogenesis and Translational Research (Ministry of Education/Beijing), Department of Radiation Oncology, Peking University Cancer Hospital \& Institute, Beijing 100142, China.}
\address{$^3$ Department of Technology, CAS Ion Medical Technology Co., Ltd., Beijing 100190, China.}
\address{$^4$ Department of Health Technology and Informatics, The Hong Kong Polytechnic University, Hong Kong SAR, 999077, China.}

\address{Jueye Zhang and Chao Yang contributed equally to this work.}
\address{ $^*$ Authors to whom any correspondence should be addressed.}

\ead{zhangyibao@pku.edu.cn}

\begin{abstract}

\textit{Purposes.} Head-and-neck cancer (HNC) treatment planning is challenging due to the close proximity of multiple critical organs-at-risk (OARs) to complex target volumes. Intensity-modulated carbon-ion therapy (IMCT) is attractive for HNC due to superior dose conformity and OAR sparing, but its planning process is slow owing to additional modeling requirements such as relative biological effectiveness (RBE). Consequently, human planners often face laborious iterative treatment-planning parameters (TPP) tuning and rely on experience-driven, suboptimal exploration. Recent studies have applied deep learning (DL) and reinforcement learning (RL) to automate treatment planning, where DL-based methods often struggle with plan feasibility and optimality due to training data bias, while RL-based methods face challenges in efficiently exploring the large and exponentially complex TPP search space. \textit{Methods.} We propose a scalable MARL framework that directly addresses these bottlenecks and enables parallel tuning of 45 TPPs for IMCT. Technically, we adopt a centralized-training decentralized-execution (CTDE) QMIX backbone augmented by Double DQN, Dueling DQN and recurrent state encoding (DRQN) to stabilize learning in a high-dimensional, non-stationary environment. To improve practicality and sample efficiency we (1) use compact historical DVH vectors as state inputs, (2) introduce a linear action-to-value transformation that maps small discrete actions to uniformly distributed parameter adjustments, and (3) design an absolute, clinically informed piecewise reward aligned to a comprehensive plan scoring system. A synchronous multi-process data-worker architecture interfaces with the PHOENIX TPS for parallel plan optimization and accelerated data collection. \textit{Results.} On a head-and-neck dataset (10 training, 10 testing) the method tuned 45 parameters simultaneously and yielded plans comparable to or better than expert manual plans (relative plan score: RL $85.93\pm7.85\%$ vs Manual $85.02\pm6.92\%$), showing significant (p-value $<$ 0.05) improvements for five OARs. \textit{Conclusions.} The results demonstrate the capability of the framework to efficiently search for high-dimensional TPPs and produce clinically competitive plans through direct TPS interaction especially for OARs.

\end{abstract}

%
\vspace{2pc}
\noindent{\it Keywords}: parallel multi-agent reinforcement learning, automatic treatment planning, intensity-modulated carbon-ion therapy, head-and-neck cancer
%
%
%
%

\section{Introduction}

Head-and-neck cancer (HNC) is a great challenge for radiotherapy: multiple critical organs-at-risk (OARs) lie close to complex target volumes, so achieving acceptable trade-offs often requires tuning many interdependent treatment-planning parameters (TPPs). Intensity-modulated carbon-ion therapy (IMCT) is attractive for HNC because its sharp dose fall-off has the potential to better protect nearby OARs \cite{kooy2015intensity,kanematsu2013treatment}. At the same time, IMCT planning brings additional modeling and delivery considerations (e.g. Relative Biological Effectiveness (RBE) and range sensitivity) that increase per-trial computational cost. In practice this makes manual tuning laborious: each parameter change typically requires a time-consuming TPS run, and planners therefore navigate the large TPP space in a experience-guided, local manner that may end up with suboptimal plans.

To reduce interactive effort and explore parameter space more systematically, many researchers have tried to explore the usage of machine / deep learning (ML / DL) for automatic treatment planning. From Dose-volume Histogram (DVH) prediction to dose map prediction \cite{wang2020review}, these methods aim to predict a potentially optimal solution to guide the optimization. However, this kind of methods rely much on the solution quality in the training dataset and produce a possible rather than feasible solution which can be directly evaluated. Meanwhile, many researchers have developed reinforcement learning (RL)-based methods \cite{shen2019intelligent, shen2020operating, shen2021improving, shen2021hierarchical, gao2023implementation, gao2024human, wang2025automating}, RL-based methods learn policies through systematic interaction with the TPS, enabling automated, data-driven exploration of the large TPP space and the generation of feasible plans without paired supervised examples. However, complex cases (e.g., HNC) require tuning a large number of TPPs, which leads to a high-dimensional action space and makes learning challenging\cite{dulac2021challenges}. As can be seen in Tab.\ref{tab:comp}, prior methods mitigate the curse of dimensionality either by tuning a small set (4 $\sim$ 5) of TPPs or dynamically choosing a subset (1 or 16) to adjust. Obviously, both of these methods are just a compromise, where still exists the problem of inefficiency (more iterations of optimization to tune all the TPPs) and suboptimal.

\begin{table}[!h]
\centering
\caption{Comparison between existing works of automatic TPP tuning.}
\label{tab:comp}
\resizebox{\textwidth}{!}{%
\begin{tabular}{llll}
\hline
Name                                       & Algorithm & Number of TPPs & Number of Parallel Tuning TPPs         \\ \hline
WTPN\cite{shen2019intelligent}, VTPN\cite{shen2020operating}, KgDRL\cite{shen2021improving}   & IQL\cite{tan1993multi}, Action Masking\cite{huang2020closer}                                                 & 4 $\sim$ 5  & 4 $\sim$ 5 \\
HieVTPN\cite{shen2021hierarchical}, VTP\cite{gao2023implementation, gao2024human} & HieDRL (Hierachical Action Space\cite{fan2019hybrid} \& SDQN\cite{metz2017discrete}) & 5 $\sim$ 48 & 1      \\
Qingqing W, et.al.\cite{wang2025automating} & PPO\cite{schulman2017proximal}       & 76             & 16            \\
Ours                                       & QMix\cite{rashid2020monotonic}      & 45             & 45  \\ \hline
\end{tabular}%
}
\end{table}

From the perspective of multi-agent reinforcement learning (MARL), this problem can be addressed with three common paradigms: centralized-training centralized-execution (CTCE; a single super-agent selects all actions), decentralized-training decentralized-execution (DTDE; independent agents act without coordination) and centralized-training decentralized-execution (CTDE; agents are trained with shared information but execute independently)\cite{gronauer2022multi}. Under such a framework, WTPN\cite{shen2019intelligent}, VTPN\cite{shen2020operating} and KgDRL\cite{shen2021improving} follow a DTDE scheme, which is straightforward but more likely to fail as the number of agents increases, due to limited global information and a more severe non-stationary environment\cite{gronauer2022multi}. Moreover, Qingqing W. et al. \cite{wang2025automating} use a CTCE scheme, which is simple to implement but suffers more from the curse of dimensionality because it ensembles all policies into a single super-agent\cite{gronauer2022multi}. Additionally, HieVTPN\cite{shen2021hierarchical} and VTP\cite{gao2023implementation, gao2024human} reduce the dimensions into one and cannot be viewed as MARL.

To balance the complexity and usage of global information, a parallel MARL algorithm based on CTDE scheme was proposed by us for automatic IMCT treatment planning, based on QMix \cite{rashid2020monotonic}, DDQN \cite{van2016deep}, Dueling DQN \cite{wang2016dueling} and DRQN \cite{hausknecht2015deep}. Additionally, the action, state and reward were novelly designed to reduce the complexity and better integrate human experience. Moreover, synchronous parallel manner \cite{liu2024acceleration} was adopted to improve the sampling efficiency in such a time-consuming environment. Eventually, our algorithm successfully worked with huge number (45) of parallel tuning TPPs and achieved better performance than human planners ("RL": $85.93 \pm 7.85\%$, "Manual": $85.02 \pm 6.92\%$) especially  for organs-at-risk (OARs). Furthermore, to the best of our knowledge, our algorithm is the first one for RL-based automatic TPP tuning in IMCT.

\section{Materials and methods}

\subsection{Plan setting}

We collected a dataset of head-and-neck treatment plans (10 for training and 10 for testing). The test cases were planned by an expert medical physicist. Clinical target volume (CTV) was selected as the target, which was extended from the Gross tumor volume (GTV) and limited to 1 mm near organs at risk (OARs) following \cite{wang2019intensity}. 18 OARs were utilized to optimize and evaluate the plans (Brainstem, Spinal Cord, Optic Chiasm, Optical-nerve Left \& Right, Temporal Lobe Left \& Right, Mandible, TMJ Left \& Right, Parotid Left \& Right, Lens  Left \& Right, Eye Left \& Right and Inner-ear Left \& Right). The treatment plans consisted of two oblique fields positioned at 45° and 315° in the transverse plane, symmetrically arranged with respect to the anterior-posterior axis of the patient. The treatment modality was carbon ion and Linear–Quadratic (LQ) served as the RBE model. Pencil beam scanning (PBS) was selected as the treatment technique. PHOENIX treatment planning system (TPS)™ (CAS Ion Medical Technology Co., Ltd., Beijing, China) was used to optimize the treatment plans.

\subsection{Plan optimization}

Treatment planning can be formulated as an optimization problem:

\begin{equation}
\begin{aligned}
\min_{f} \quad & \sum_{m}^M w_m \, \mathcal{O}_m(f; v_m) \\
\text{s.t.} \quad & f \geq 0
\end{aligned}
\end{equation}

It is a kind of multi-objective optimization problem. Several ($M$) clinical objectives $\mathcal{O}_m$ are used to find the best fluence map $f$ inversely. $w_m$ and $v_m$ are the weight and the hyper-parameter (e.g., dose threshold, volume threshold) of each clinical objective respectively. The detailed clinical objective list can be found in the Tab.\ref{tab:parameter_list} in \ref{app:A}. 

\subsection{Plan parameters tuning via parallel multi-agent reinforcement learning}

\begin{figure}[!h]
    \centering
    \includegraphics[width=1\linewidth]{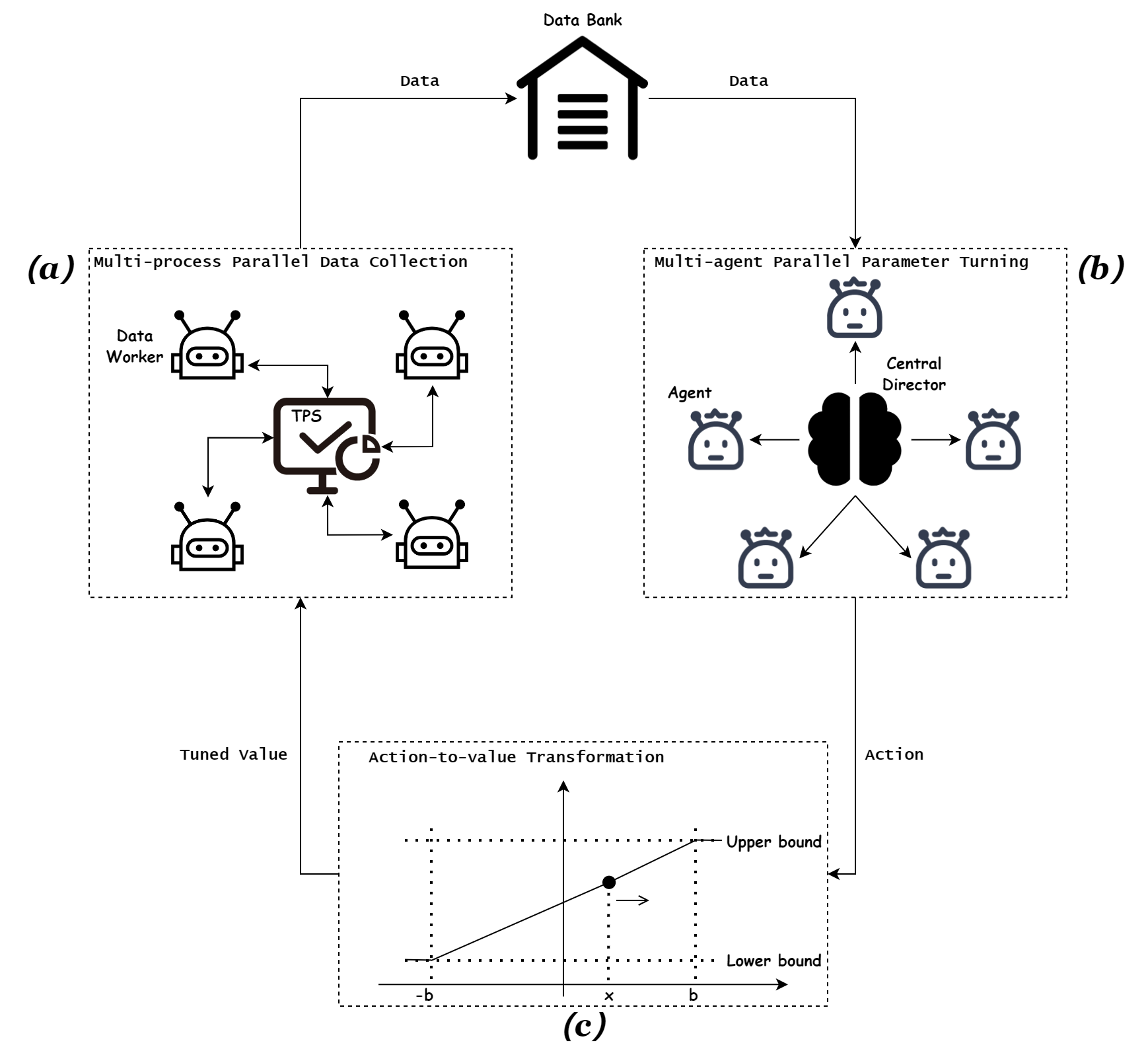}
    \caption{The overview of our parallel MARL algorithm. \textbf{\textit{(a)}} Multi-process data workers interact with TPS in parallel to fill the data bank with planning experience. \textbf{\textit{(b)}} The multi-agent system retrieves data from the data bank to optimize its policy under the supervision of a central director, who facilitates global communication among different agents. \textbf{\textit{(c)}} The trained policy subsequently guides data collection through an action-to-value transformation. Data collection and agent training are carried out sequentially and mutually enhance each other, ultimately resulting in high-quality plan data in the data bank and a well-trained treatment planner.}
    \label{fig:method}
\end{figure}

Our design choices target two interrelated problems: (1) algorithmic instability from many interacting learners, and (2) practical constraints of costly TPS interactions.
Firstly, to address algorithmic instability we employ a CTDE value-decomposition backbone (QMIX) so that agents can learn with access to global information during training while preserving decentralized execution. QMIX enforces positive monotonicity to simplify the mapping from per-agent Q-values to a global value. Secondly, to reduce overestimation and improve robustness we integrate Double DQN and Dueling DQN, and we use recurrent state encoding (DRQN) to exploit historical DVH trajectories for temporally coherent decisions. Thirdly, to make actions clinically meaningful and uniformly resolvable across parameters, we adopt a linear action-to-value transformation that maps small discrete actions into uniformly spaced parameter values within human-informed bounds, to avoid the exponential density bias in previous exponential mappings. Finally, recognizing the high cost of each TPS call, we engineered a synchronous multi-process data-worker system to run multiple plan optimizations in parallel against the PHOENIX TPS, improving empirical sample throughput while preserving realistic evaluation. The technical details are described as follows.

We modeled the treatment plan parameters tuning as a Multi-agent Markov Decision Process defined by the tuple $( \mathcal{S}, \{\mathcal{A}^i\}_{i=1}^N, \mathcal{P}, \mathcal{R}, \gamma )$, where:

\begin{figure}[!h]
    \centering
    \includegraphics[width=1\linewidth]{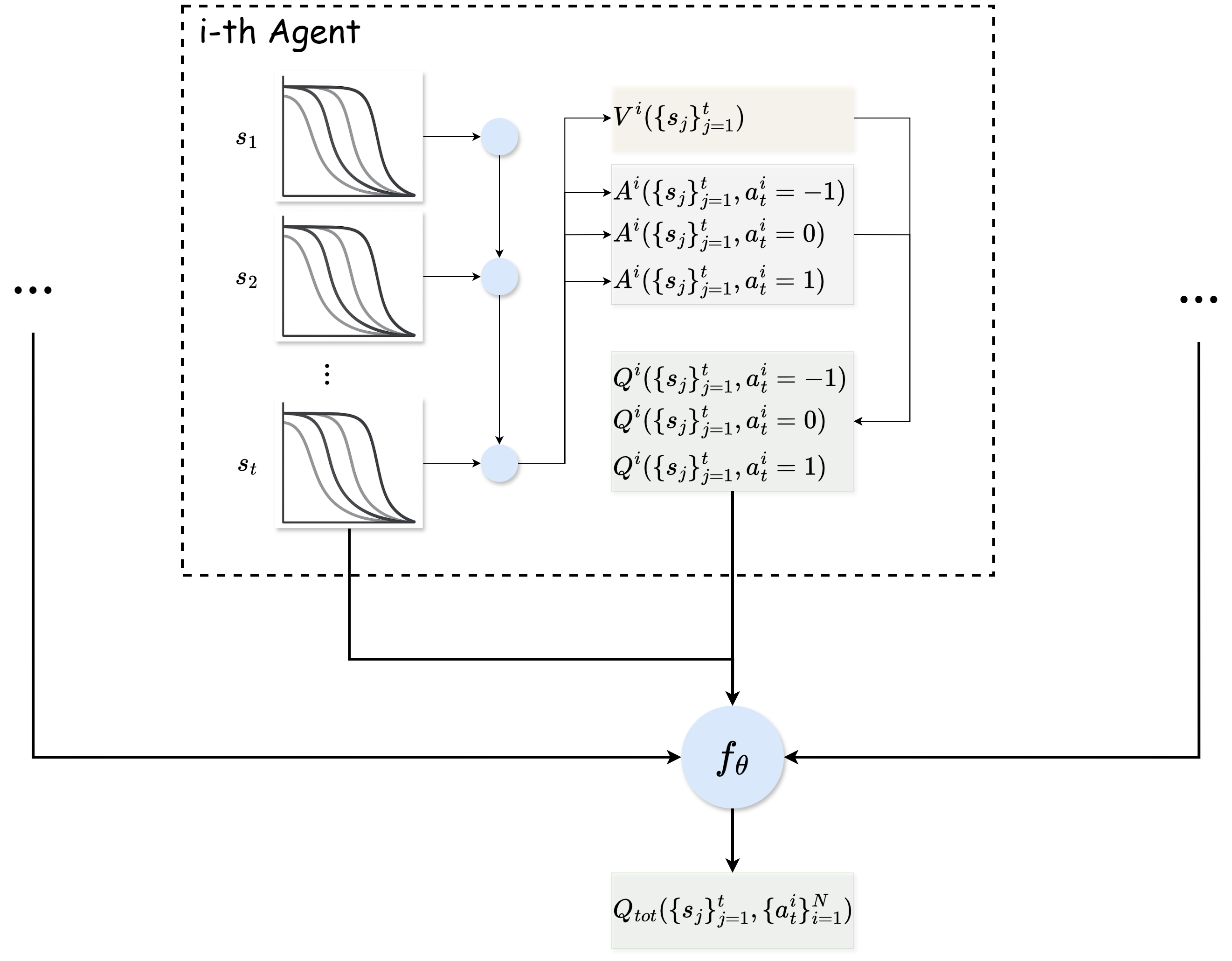}
    \caption{The overview of our multi-agent system. Every agent take all the historical DVHs as input to estimate $V$-value and $A$-value for $Q$-value. All the estimated $Q$-value is then fed into a central director $f_{\theta }$ to get the final estimation of $Q_{tot}$-value.}
    \label{fig:agent}
\end{figure}

\begin{itemize}
    \item \textbf{State Space} $\mathcal{S}$: Each $s_t \in \mathcal{S}$ denotes the optimization result of the current ($t \in [1,T]$, where $T$ is the episode length) parameter combination. For computational efficiency, the Dose-volume Histograms (DVHs) were adopted, instead of the whole 3-dimensional dose map. To put it another way, $s_t \in \mathbb{R}^{m\times n}$, where $m$ denotes the number of organs (CTV and OARs) and $n$ denotes the DVH size. In our study, $m$ was set to 19 and $n$ was set to 150.

    \item \textbf{Action Space} $\{\mathcal{A}^i\}_{i=1}^N$: $N$ is the number of agents. Each agent controls one single optimization parameter (value or weight of optimization objectives of CTV and 18 OARs). In our study, $N$ was set to 45. The detailed parameter list can be found in the Tab.\ref{tab:parameter_list} in \ref{app:A}. Each $a^i_t \in \mathcal{A}^i$ represents the changing of the $i$th parameter by the $i$th agent. To reduce the action space, the upper and lower bounds of each parameter were set based on clinical experience. Moreover, the scope between the two bounds is transferred from continuous to discrete through "action-to-value transformation", just as Fig.\ref{fig:method} shows. To be more specific, $\text{Tuned Value}^i_t  = \begin{cases} \text{Upper Bound}^i, x^i_t \ge b \\ \text{Lower Bound}^i +\frac{x^i_t+b}{2b}(\text{Upper Bound}^i-\text{Lower Bound}^i),-b<x^i_t<b\\\text{Lower Bound}^i,x^i_t\le-b \end{cases}$ and $x^i_t=x^i_{t-1}+a^i_t$, where $x^i_t$ serves as the current ($t$) x-coordinate of $i$th parameter and $a^i_t \in \{-1, 0,1\}$ (move left, stay still, move right). $b$ is a symbol of tuning resolution and was set to 5 in our study. All the $x_0^i$ were initialized to 0 and $\text{Tuned Value}^i_0=0.5 \times(\text{Upper Bound}^i-\text{Lower Bound}^i)$. Our action space was designed using a linear rather than an exponential mapping (as frequently used in prior studies \cite{shen2021hierarchical,shen2021improving}) to avoid uneven tuning density in the action space (e.g., \{$0$, ..., $e^{-0.5},e^0,e^{0.5}$, ..., $+\infty$\}, which are much denser below $e^0$, resulting in fine-grained adjustments for decreasing and coarse ones for increasing parameter values) . 

    \item \textbf{State transition function} $\mathcal{P}$: $\mathcal{P}(s_{t+1}|s_{t},\{a^i_t\}_{i=1}^N)$ denotes the transition function of DVHs with different groups of optimization parameters, which is based on the optimization engine of TPS.

    \item \textbf{Reward Function} $\mathcal{R}$: Each $r_t(s_t,\{a_t\}_{i=1}^N) \in \mathcal{R}$ stands for the immediate reward got if $\{a_t\}_{i=1}^N$ are taken at state $s_t$. To lay the groundwork, we developed a scoring system (defined as $\phi(s_t)$, shown in the Tab.\ref{tab:scoring_criteria} in \ref{app:A}) according to the former study \cite{hu2021quantitative} and treatment guidelines \cite{CACA_particle_therapy_2024}, to evaluate the plan quality. The total plan score was calculated through summing up the scores of the CTV and 18 OARs. Thresholds of each OAR metric were based on the dose restrictions \cite{CACA_particle_therapy_2024} and scores of each OAR \& CTV metric were based on the organ importance \cite{hu2021quantitative}. The reward was set to $r_t=\phi(s_{t+1})-0.5\times max\phi(*)$, in an absolute rather than frequently used relative form \cite{shen2021hierarchical, gao2024human, madondo2025patient} (i.e., $r_t'=\phi(s_{t+1})- \phi(s_{t})$). The specific design was for higher reward assigned with same improvement at the higher score scope (e.g., if $\phi(s_t)=40, \phi(s_{t+1})=50,\phi(s_{t+2})=60$ then $r_{t+1} > r_t$ while $r_t'=r_{t+1}'$). The constant term $0.5\times max\phi(*)$ was used to avoid ultra-high positive reward.

    \item \textbf{Discount factor} $\gamma$: $\gamma$ is the discount factor applied to future rewards. In our study, it was set to $0.9 \in (0,1)$.
\end{itemize}

\begin{figure}[!h]
    \centering
    \includegraphics[width=1\linewidth]{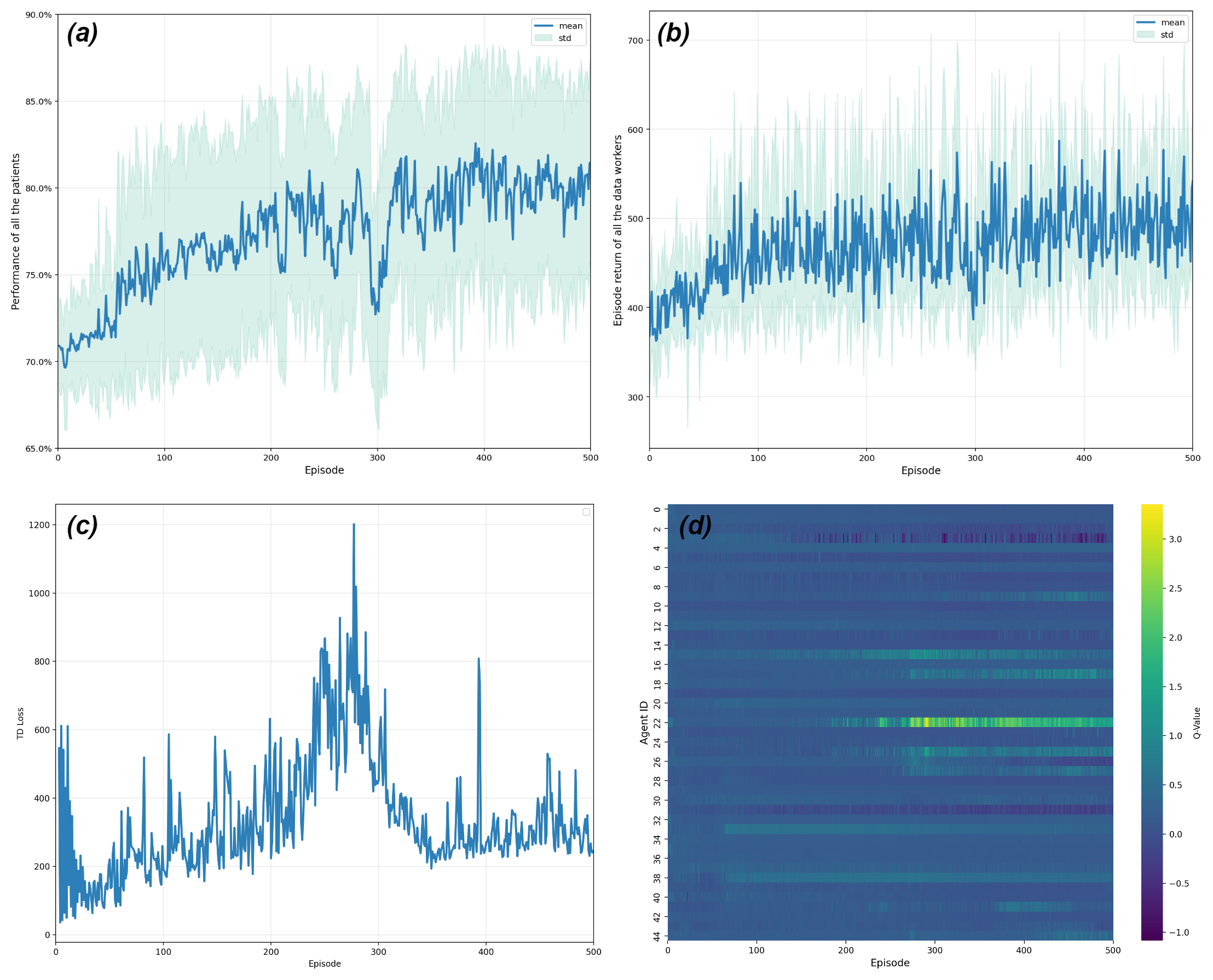}
    \caption{Training results. \textbf{\textit{(a)}} Performance of all the patients (mean $\pm$ std; shaded area indicates one standard deviation). \textbf{\textit{(b)}} Episode return of all the data workers (mean $\pm$ std; shaded area indicates one standard deviation). \textbf{\textit{(c)}} Temporal Difference (TD) loss. \textbf{\textit{(d)}} The Q-value of all the agents.}
    \label{fig:training}
\end{figure}

\begin{figure}[!t]
    \centering
    \includegraphics[width=1\linewidth]{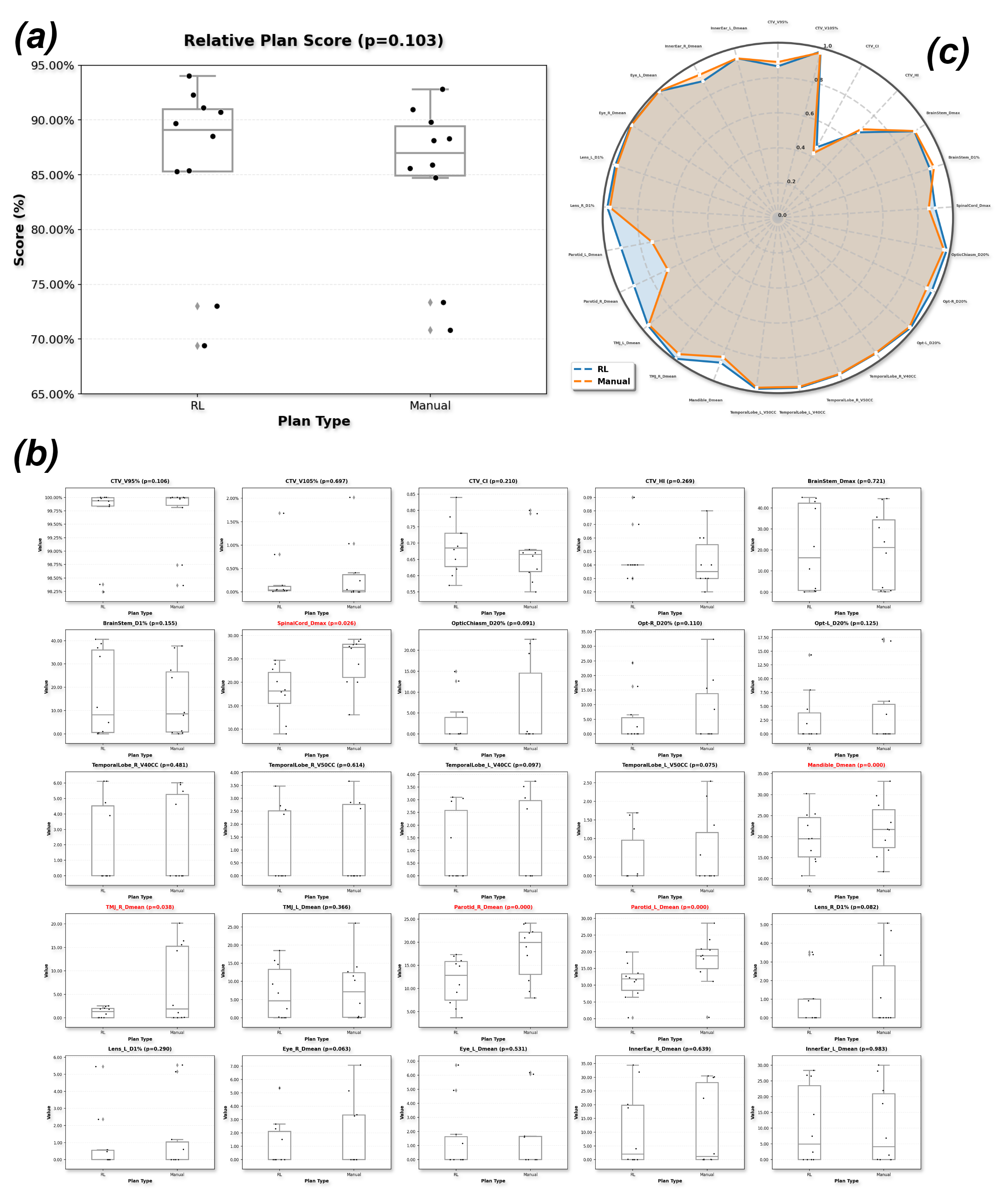}
    \caption{Statistical comparison between "RL" and "Manual". \textbf{\textit{(a)}} Box plot of the relative plan score (with p-values indicating whether "RL" outperforms "Manual"). \textbf{\textit{(b)}} Box plot of all the plan metrics (with p-values indicating whether "RL" outperforms "Manual"; highlighted in red if statistically significant at the 0.05 threshold). \textbf{\textit{(c)}} Radar chart of relative score of all the plan metrics.} 
    \label{fig:compare}
\end{figure}

\begin{figure}[!h]
    \centering
    \includegraphics[width=1\linewidth]{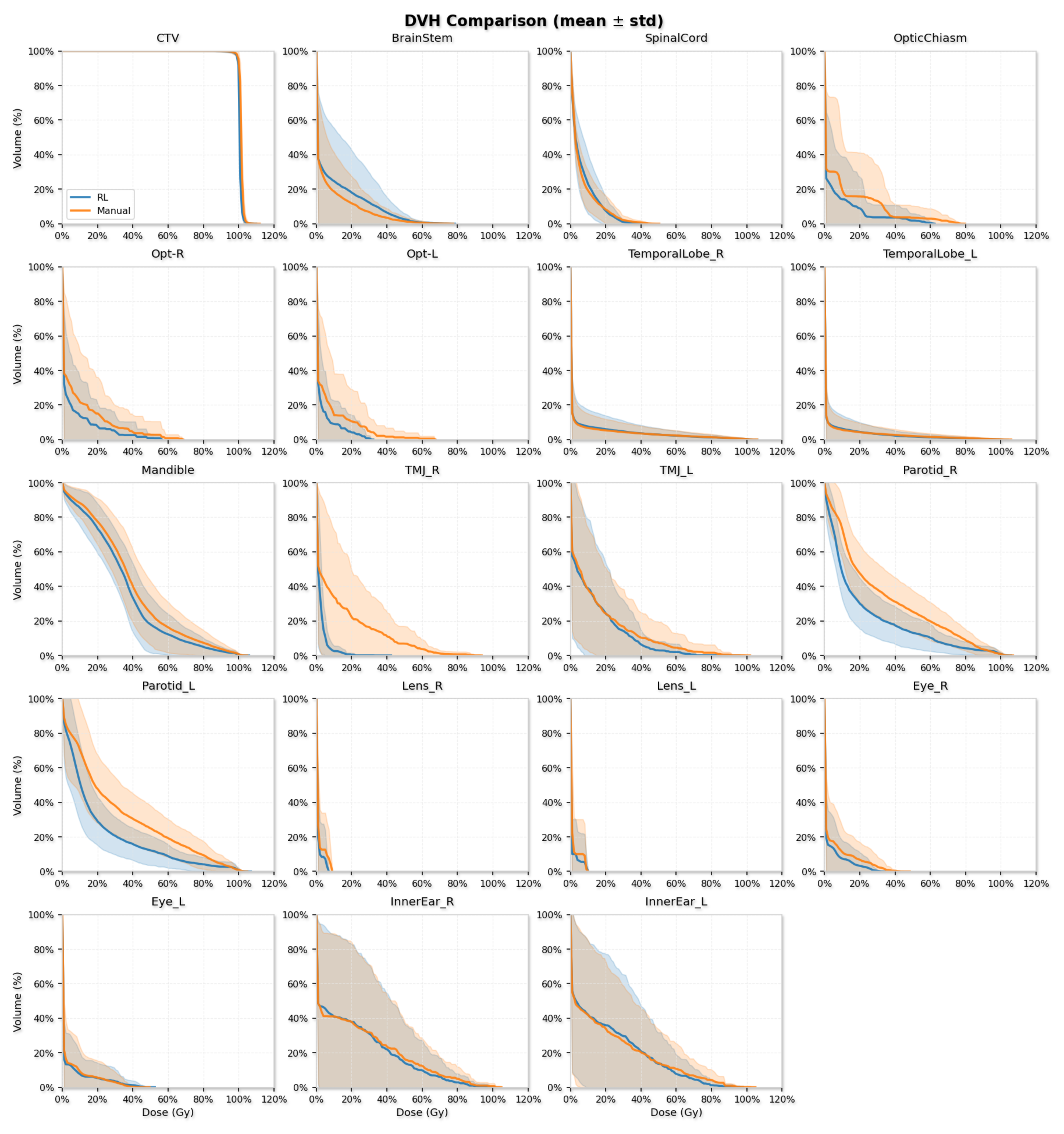}
    \caption{Average DVHs of all the OARs and CTV. (mean $\pm$ std; shaded area indicates one standard deviation)}
    \label{fig:dvh}
\end{figure}

Based on the MDP formulation above, the reinforcement learning (RL) was adopted to find the best policy ($\pi^*$). To improve the efficiency of tuning, the multi-agent reinforcement learning (MARL) was employed, which supports parallel tuning of all the parameters. Due to the discrete action space, value-based algorithm was adopted. In our value-based MARL, $Q_{tot}(\{s_j\}_{j=1}^t, \{a^i_t\}_{i=1}^N)$ was defined, which denotes the total state-action value function for potential return estimating of the state-action pair ($\{s_j\}_{j=1}^t$ and $ \{a^i_t\}_{i=1}^N$). To train every individual value function ($Q_i(\{s_j\}_{j=1}^t,a^i_t)$) estimator, the value decomposition \cite{sunehag2017value} manner was utilized. To learn the complex relationship between $Q_{tot}$ and every single $Q_i$, QMix \cite{rashid2020monotonic} was used. 

\begin{equation}
    Q_{tot}(\{s_j\}_{j=1}^t,\{a^i_t\}_{i=1}^N)=f_\theta(Q_1(\{s_j\}_{j=1}^t,a_t^1),...,Q_N(\{s_j\}_{j=1}^t,a_t^N),s_t), \frac{\partial Q_{tot}}{Q_i} \ge0
    \label{eq:qmix}
\end{equation}

In QMix \cite{rashid2020monotonic}, all the single $Q_i$ are fed into a neuro-network (NN) $f_{\theta}$ to get the final $Q_{tot}$, just as Eq.\ref{eq:qmix} shows. To avoid learning an overly complex relationship, each \(Q_i\) is constrained to be positively correlated with \(Q_{tot}\); this is achieved using hypernetworks \cite{ha2016hypernetworks}.

The design of individual estimator $Q_i$ can be seen in Fig.\ref{fig:agent}.

\textbf{DRQN}: All the past states ($\{s_j\}_{j=1}^t$) serve as an input with a sequence processing module (Long short-term memory, LSTM \cite{hochreiter1997long}, following DRQN\cite{hausknecht2015deep}) to support better decision-making, just as a medical physicist tuning parameters with knowledge of all the historical DVHs. Additionally, the optimization of TPS is time-dependent because the initialization of this optimization step is related to results of all the former steps. Thus, the agents require former states for a better decision.

Moreover, from the view of every single agent, the environment is non-stationary \cite{hernandez2019survey}, because the optimization result is based on the all the ever-changing agents. Consequently, the overestimation problem of value-based RL becomes severe \cite{pan2021regularized}, especially for huge numbers of agents. To address the problem, our method combines DDQN\cite{van2016deep} and Dueling DQN\cite{wang2016dueling}.

\begin{equation}
    Q^i(\{s_j\}_{j=1}^t,a_t^i) = V^i(\{s_j\}_{j=1}^t) + A(\{s_j\}_{j=1}^t,a_t^i) - \frac{1}{|\mathcal{A}^i|}\sum_{a'\in\mathcal{A}^i}A(\{s_j\}_{j=1}^t,a')
    \label{eq:dueling}
\end{equation}

\textbf{Dueling DQN}: As can be seen in Eq.\ref{eq:dueling},  every single $Q^i$ is divided into two parts $V^i(\{s_j\}_{j=1}^t)$ and $A(\{s_j\}_{j=1}^t,a_t^i) - \frac{1}{|\mathcal{A}^i|}\sum_{a'\in\mathcal{A}^i}A(\{s_j\}_{j=1}^t,a')$. The first part is the state-value function and $A(\{s_j\}_{j=1}^t,a_t^i)$ is the advantage function (minus $ \frac{1}{|\mathcal{A}^i|}\sum_{a'\in\mathcal{A}^i}A(\{s_j\}_{j=1}^t,a')$ to keep the sum of this function equal to zero). 

\begin{equation}
    \mathcal{L} = ||y-Q_{tot}(\{s_j\}_{j=1}^t,\{a^i_t\}_{i=1}^N)||^2, y=r_t+\gamma Q_{tot}^{tar}(\{s_j\}_{j=1}^{t+1},\{argmax_{a'}Q_i(\{s_j\}_{j=1}^{t+1},a')\}_{i=1}^N))
    \label{eq:double}
\end{equation}

\textbf{DDQN}: The loss function is shown in Eq.\ref{eq:double}. To reduce Q-value overestimation, $Q_{tot}^{tar}$ is used, which is called "target network" (following DDQN). "target network" shares the same structure with the initial one while uses a delayed update mechanism to reduce overestimation. 

A major challenge of RL in TPS is its low sampling efficiency (each step requires solving fluence map optimization, FMO \cite{long2015optimization}), compared with standard RL benchmarks. To enhance the sampling efficiency, modification of the TPS was made to support parallel optimization of different treatment plans. Just as Fig.\ref{fig:method} shows, multiple data workers were used during training to interact with the TPS for parallel data collection. Moreover, the data workers and agents operate synchronously\cite{liu2024acceleration}, which means that the data collection and model update are performed in a sequential fashion.

\subsection{Implementation details}
10 data workers were used to interact with TPS. $\epsilon$-greedy algorithm was used to select action and $\epsilon$ was set to 0.9 with a decay of 90\% every 5 episodes. The episode length ($T$) was set to 10 for truncation. No terminal state was defined because a higher plan score is always preferable. The model was trained for 15 days (500 episodes) with 2 NVIDIA RTX A6000 GPUs.

\section{Results}

\subsection{Training results.}

The training results can be found in Fig.\ref{fig:training}. Fig.\ref{fig:training}.a shows the performance (max relative plan score in an episode $\frac{max_{t\in [1,T]}\phi(s_t)}{max \phi(*)}$) over all the training cases (mean and std) in the process of training. Fig.\ref{fig:training}.b is the episode return ($\sum_{t=1}^Tr_t$) over all the data workers (mean and std). Both the average performance and average episode return curves rise along with the training process. Fig.\ref{fig:training}.C shows the Temporal Difference (TD) loss defined in Eq.\ref{eq:qmix} and Fig.\ref{fig:training}.D plots the Q-value of all the agents reflecting the policy evolution. As is shown in Fig.\ref{fig:training}.C and Fig.\ref{fig:training}.D, a significant policy change occurred between episodes 200 and 300, after which the policy gradually stabilized.

\subsection{Comparison results between "RL" and "Manual".}

Overall comparisons on the testing set are shown in Figs.\ref{fig:compare} and \ref{fig:dvh}. Fig.\ref{fig:compare}a presents box plots of the relative plan score (RL: $85.93 \pm 7.85\%$, Manual: $85.02 \pm 6.92\%$). Fig.\ref{fig:compare}.b gives box plots for each plan metric listed in Table\ref{tab:scoring_criteria} (Appendix\ref{app:A}), and Fig.\ref{fig:compare}.c displays a radar chart of the metric-wise relative scores (RL in blue, Manual in red). Fig.\ref{fig:dvh} shows the mean DVHs for CTV and OARs, with shaded areas indicating $\pm$ standard deviation.

Overall, RL plans perform comparably to manual plans for the CTV and most OARs, and they show clear improvements for five OARs: SpinalCord, Mandible, TMJ\_R, Parotid\_R and Parotid\_L. The advantage for the parotids is particularly evident in the radar chart. The average DVHs in Fig.\ref{fig:dvh} indicate that, for similar CTV coverage, the RL plans yield lower doses to almost all OARs except the brainstem.


\begin{figure}[!h]
    \centering
    \includegraphics[width=1\linewidth]{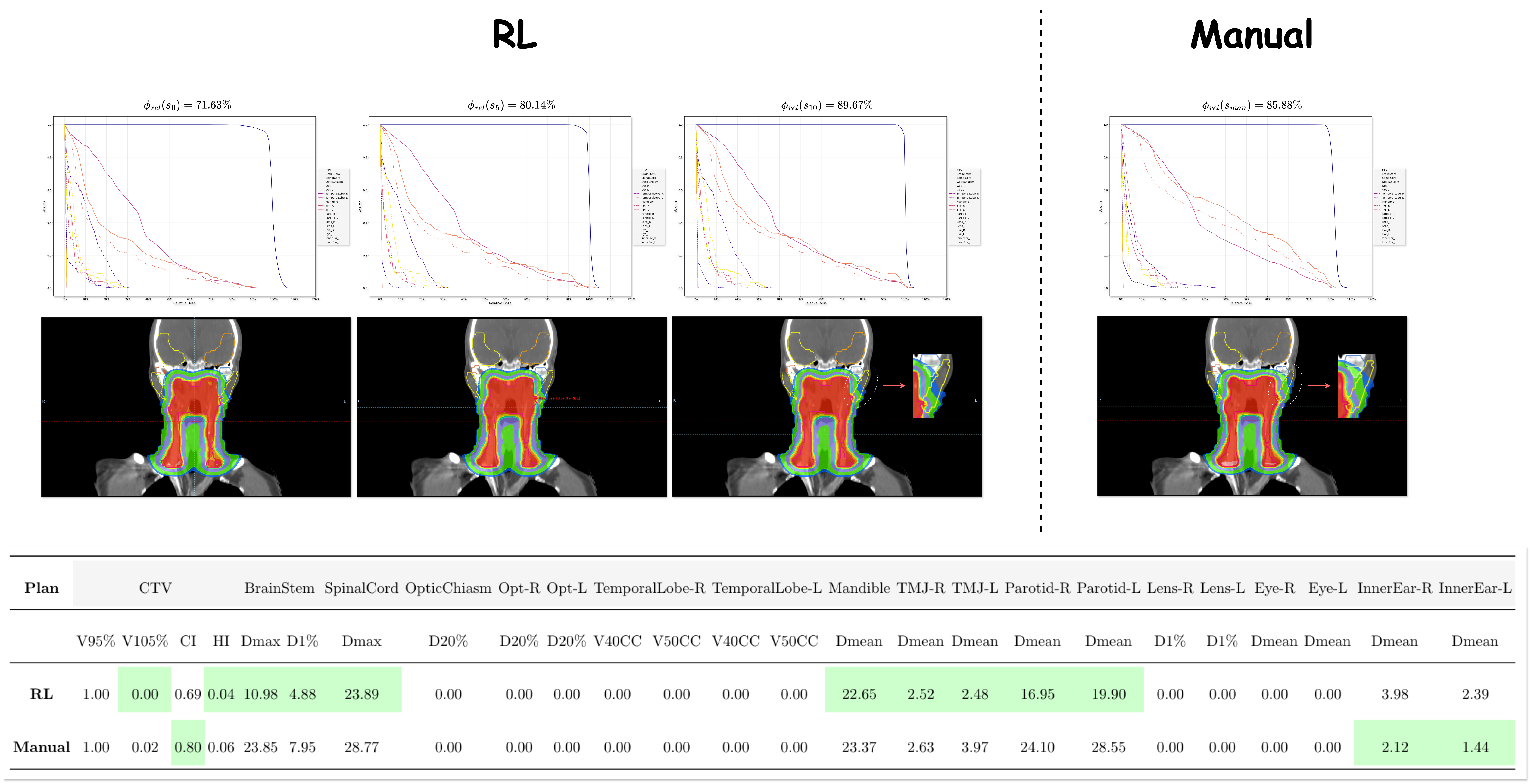}
    \caption{One case for episode results visualization.}
    \label{fig:episode}
\end{figure}

To show the progress of plan change in the episode, Fig.\ref{fig:episode} is presented, in which a case is selected to show the change of both DVH and dose map. The relative plan score keeps rising from $71.63\%$ at initialization to $80.14\%$ at middle stage and eventually achieves $89.67\%$. From the DVHs, the performance of CTV becomes better step by step, leading to unavoidable sacrifice to some OARs. To be compare, the manual plan achieves a relative score of $85.88\%$. In the dose map, better protection of several OARs is shown, especially the parotid.

\section{Discussions}

Due to the exponential increase of action space size along with the number of TPPs ($3^N$, $3$: action space size for a single agent, $N$: number of agents / parallel tuning TPPs), it seems impossible for RL to work well in the scenario of complex cases with lots of OARs without special design, especially for head \& neck cases. Thus, former studies tend to employ the strategy of reducing number of parallel tuning TPPs to varying degrees. To the best of our knowledge, Qingqing W, et,al.\cite{wang2025automating} achieves the most number (16) of parallel tuning TPPs, with dynamically TPP selection. From the MARL perspective, previous studies have focused on DTDE and CTCE. DTDE scheme allows agents to learn dependently without information sharing, thus facing great challenge in a complex environment with numerous independent learners\cite{gronauer2022multi}. CTCE scheme employs a super-agent to conduct all the decisions and suffers more by curse of dimensionality \cite{gronauer2022multi}. To strike a balance between centralization and decentralization, this study proposed a parallel CTDE algorithm for large-scale TPP tuning in automatic IMCT. Apart from the MARL learning algorithm, state (including all the historical DVHs for better decision with comprehensive information), action (linear interpolation via human-set upper and lower bound for tuning uniformity), reward (absolute rather than relative for better reward assignment), environment (parallel optimization and easy to scale-up for more stable training), learning target (DDQN \& Dueling DQN for stability) were also carefully designed to work well in a complex scenario with large-scale TPPs (45) to be parallel tuned.

As is shown in Fig.\ref{fig:training}, both the performance and return rise along the training process. Moreover, the TD loss and Q-values gradually become stable after a shape curve between 200th and 300th episode. The above results demonstrate the stability of our method to learn in such a complex environment. To demonstrate the ability of our method comprehensively, a detailed analysis has been conducted in the testing dataset. Fig.\ref{fig:compare} is consist of box plot of all the planning metrics and radar chat for more intuitive comparison, where "RL" outperforms "Manual" in 5 OARs significantly. Fig.\ref{fig:dvh} is the figure of mean DVHs between the two types of plans. As is presented, "RL" achieved a comparable or even better performance than “Manual”, especially in the protection of OARs. Only the "RL" DVH of the brainstem appears worse than the "Manual" one, but given that the brainstem is a type of serial organ that does not tolerate local high doses (as reflected by $D_{max}/D_{1\%}$), this difference is clinically negligible. The result may be attributed to the reward design based on the plan scoring criteria, which is introduced in Tab.\ref{tab:scoring_criteria} in \ref{app:A}. Further but slight credits were provided for the agent to pursue even reaching the clinical goals for fully idealization ($\rightarrow 0$). Fine tuning TPPs in these regimes may be prohibitively complex for human planners, but feasible for RL agents with high-throughput, parallel tuning. Fig.\ref{fig:episode} shows the episode plan change of one case, where RL algorithm do trade-offs between OARs and CTV to improve the plan score.

Despite these encouraging results, several limitations remain. The current study uses a relatively small dataset and a single TPS (PHOENIX). Broader validation across more cases, tumor sites and planning systems is needed. Future work will focus on scaling to larger datasets, reducing sample complexity via model-based or offline RL techniques, and performing multi-institutional studies to evaluate generalizability and workflow integration.

\section{Conclusion}
We have presented a parallel multi-agent reinforcement learning approach for large-scale automatic tuning of TPPs in IMCT for HNC within an extensive search space. By modeling each TPP as an agent and employing a CTDE scheme with QMIX, DRQN, Double DQN and Dueling DQN, the proposed framework is able to tune 45 parameters in parallel, leverage historical DVH information, and learn stable policies in a time-consuming TPS environment via synchronous multi-process data collection. On the testing dataset, our method produced plans of comparable or superior quality to expert manual plans, particularly improving sparing for several OARs.

\bibliographystyle{./unsrt.bst}    

\bibliography{./main.bbl}      

\appendix

\section{Supplementary Tables}
\label{app:A}

\begin{table}[!h]
\centering
\caption{Parameter list of automatic treatment planning}
\label{tab:parameter_list}

\renewcommand{\arraystretch}{0.8} 
\begin{tabular}{lll}
\toprule
\textbf{Organ} & \textbf{Objective} & \textbf{Parameters} \\
\midrule
\multirow{4}{*}{CTV} 
    & Max       & obj\_value, weight \\
    & Min       & obj\_value, weight \\
    & Uniform   & obj\_value, weight \\
    & DVHmin    & Dose, Volume, weight \\
\midrule
BrainStem           & Max       & obj\_value, weight \\
SpinalCord          & Max       & obj\_value, weight \\
OpticChiasm         & Max       & obj\_value, weight \\
Opt-R               & Max       & obj\_value, weight \\
Opt-L               & Max       & obj\_value, weight \\
TemporalLobe\_R      & Max       & obj\_value, weight \\
TemporalLobe\_L      & Max       & obj\_value, weight \\
Mandible            & Max       & obj\_value, weight \\
TMJ\_R              & Max       & obj\_value, weight \\
TMJ\_L              & Max       & obj\_value, weight \\
Parotid\_R          & Max       & obj\_value, weight \\
Parotid\_L          & Max       & obj\_value, weight \\
Lens\_R             & Max       & obj\_value, weight \\
Lens\_L             & Max       & obj\_value, weight \\
Eye\_R              & Max       & obj\_value, weight \\
Eye\_L              & Max       & obj\_value, weight \\
InnerEar\_R         & Max       & obj\_value, weight \\
InnerEar\_L         & Max       & obj\_value, weight \\
\bottomrule
\end{tabular}
\end{table}

{\small
\renewcommand{\arraystretch}{0.4}
\begin{longtable}{ll}

\caption{Radiotherapy Plan Scoring Criteria. The score for each metric is calculated based on its value, denoted as $v$ in the formulas.}
\label{tab:scoring_criteria} \\

\toprule
\textbf{Quantity of Interest} & \textbf{Scoring Criterion (where $v$ is the value)} \\
\midrule
\endfirsthead

\multicolumn{2}{l}{Table \thetable{} -- Continued} \\
\toprule
\textbf{Quantity of Interest} & \textbf{Scoring Criterion (where $v$ is the value)} \\
\midrule
\endhead

\midrule
\multicolumn{2}{r}{\textit{Continued on next page}} \\
\endfoot

\bottomrule
\endlastfoot

CTV $V_{95\%}$ \text{(\%)} &
$ \text{Score} = \Biggl\lbrace \renewcommand{\arraystretch}{0.7}\begin{array}{@{}l@{\quad}l@{}}
-40 & \text{if } v \le 0 \\
-40 + \frac{52}{0.98}v & \text{if } 0 < v \le 0.98 \\
12 + \frac{28}{0.02}(v - 0.98) & \text{if } 0.98 < v \le 1 \\
40 & \text{if } v > 1
\end{array} $ \\\\
CTV $V_{105\%}$ \text{(\%)} &
$ \text{Score} = \Biggl\lbrace \renewcommand{\arraystretch}{0.7}\begin{array}{@{}l@{\quad}l@{}}
20 & \text{if } v \le 0 \\
20 - \frac{14}{0.1}v & \text{if } 0 < v \le 0.1 \\
6 - \frac{6}{0.9}(v - 0.1) & \text{if } 0.1 < v \le 1 \\
0 & \text{if } v > 1
\end{array} $ \\\\
CTV $\text{CI}$ &
$ \text{Score} = \Biggl\lbrace \renewcommand{\arraystretch}{0.7}\begin{array}{@{}l@{\quad}l@{}}
0 & \text{if } v \le 0 \\
\frac{6}{0.6}v & \text{if } 0 < v \le 0.6 \\
6 + \frac{14}{0.4}(v - 0.6) & \text{if } 0.6 < v \le 1 \\
20 & \text{if } v > 1
\end{array} $ \\\\
CTV $\text{HI}$ &
$ \text{Score} = \Biggl\lbrace \renewcommand{\arraystretch}{0.7}\begin{array}{@{}l@{\quad}l@{}}
20 & \text{if } v \le 0 \\
20 - \frac{14}{0.1}v & \text{if } 0 < v \le 0.1 \\
6 - \frac{6}{0.9}(v - 0.1) & \text{if } 0.1 < v \le 1 \\
0 & \text{if } v > 1
\end{array} $ \\\\
BrainStem $D_{\text{max}}$ \text{(Gy)} &
$ \text{Score} = \Biggl\lbrace \renewcommand{\arraystretch}{0.7}\begin{array}{@{}l@{\quad}l@{}}
7.2 & \text{if } v \le 0 \\
7.2 - \frac{1.2}{45}v & \text{if } 0 < v \le 45 \\
6 - \frac{6}{9}(v - 45) & \text{if } 45 < v \le 54 \\
0 & \text{if } v > 54
\end{array} $ \\\\
BrainStem $D_{1\%}$ \text{(Gy)} &
$ \text{Score} = \Biggl\lbrace \renewcommand{\arraystretch}{0.7}\begin{array}{@{}l@{\quad}l@{}}
7.2 & \text{if } v \le 0 \\
7.2 - \frac{1.2}{38.5}v & \text{if } 0 < v \le 38.5 \\
6 - \frac{6}{11.5}(v - 38.5) & \text{if } 38.5 < v \le 50 \\
0 & \text{if } v > 50
\end{array} $ \\\\
SpinalCord $D_{\text{max}}$ \text{(Gy)} &
$ \text{Score} = \Biggl\lbrace \renewcommand{\arraystretch}{0.7}\begin{array}{@{}l@{\quad}l@{}}
14.4 & \text{if } v \le 0 \\
14.4 - \frac{2.4}{30}v & \text{if } 0 < v \le 30 \\
12 - \frac{12}{10}(v - 30) & \text{if } 30 < v \le 40 \\
0 & \text{if } v > 40
\end{array} $ \\\\
OpticChiasm $D_{20\%}$ \text{(Gy)} &
$ \text{Score} = \Biggl\lbrace \renewcommand{\arraystretch}{0.7}\begin{array}{@{}l@{\quad}l@{}}
14.4 & \text{if } v \le 0 \\
14.4 - \frac{2.4}{30}v & \text{if } 0 < v \le 30 \\
12 - \frac{12}{10}(v - 30) & \text{if } 30 < v \le 40 \\
0 & \text{if } v > 40
\end{array} $ \\\\
Opt-R $D_{20\%}$ \text{(Gy)} &
$ \text{Score} = \Biggl\lbrace \renewcommand{\arraystretch}{0.7}\begin{array}{@{}l@{\quad}l@{}}
7.2 & \text{if } v \le 0 \\
7.2 - \frac{1.2}{30}v & \text{if } 0 < v \le 30 \\
6 - \frac{6}{10}(v - 30) & \text{if } 30 < v \le 40 \\
0 & \text{if } v > 40
\end{array} $ \\\\
Opt-L $D_{20\%}$ \text{(Gy)} &
$ \text{Score} = \Biggl\lbrace \renewcommand{\arraystretch}{0.7}\begin{array}{@{}l@{\quad}l@{}}
7.2 & \text{if } v \le 0 \\
7.2 - \frac{1.2}{30}v & \text{if } 0 < v \le 30 \\
6 - \frac{6}{10}(v - 30) & \text{if } 30 < v \le 40 \\
0 & \text{if } v > 40
\end{array} $ \\\\
TemporalLobe-R $V_{40\text{cc}}$ \text{(cc)} &
$ \text{Score} = \Biggl\lbrace \renewcommand{\arraystretch}{0.7}\begin{array}{@{}l@{\quad}l@{}}
3.6 & \text{if } v \le 0 \\
3.6 - \frac{0.6}{7.66}v & \text{if } 0 < v \le 7.66 \\
3 - \frac{3}{2.34}(v - 7.66) & \text{if } 7.66 < v \le 10 \\
0 & \text{if } v > 10
\end{array} $ \\\\
TemporalLobe-R $V_{50\text{cc}}$ \text{(cc)} &
$ \text{Score} = \Biggl\lbrace \renewcommand{\arraystretch}{0.7}\begin{array}{@{}l@{\quad}l@{}}
3.6 & \text{if } v \le 0 \\
3.6 - \frac{0.6}{4.66}v & \text{if } 0 < v \le 4.66 \\
3 - \frac{3}{1.34}(v - 4.66) & \text{if } 4.66 < v \le 6 \\
0 & \text{if } v > 6
\end{array} $ \\\\
TemporalLobe-L $V_{40\text{cc}}$ \text{(cc)} &
$ \text{Score} = \Biggl\lbrace \renewcommand{\arraystretch}{0.7}\begin{array}{@{}l@{\quad}l@{}}
3.6 & \text{if } v \le 0 \\
3.6 - \frac{0.6}{7.66}v & \text{if } 0 < v \le 7.66 \\
3 - \frac{3}{2.34}(v - 7.66) & \text{if } 7.66 < v \le 10 \\
0 & \text{if } v > 10
\end{array} $ \\\\
TemporalLobe-L $V_{50\text{cc}}$ \text{(cc)} &
$ \text{Score} = \Biggl\lbrace \renewcommand{\arraystretch}{0.7}\begin{array}{@{}l@{\quad}l@{}}
3.6 & \text{if } v \le 0 \\
3.6 - \frac{0.6}{4.66}v & \text{if } 0 < v \le 4.66 \\
3 - \frac{3}{1.34}(v - 4.66) & \text{if } 4.66 < v \le 6 \\
0 & \text{if } v > 6
\end{array} $ \\\\
Mandible $D_{\text{mean}}$ \text{(Gy)} &
$ \text{Score} = \Biggl\lbrace \renewcommand{\arraystretch}{0.7}\begin{array}{@{}l@{\quad}l@{}}
6 & \text{if } v \le 0 \\
6 - \frac{1}{30}v & \text{if } 0 < v \le 30 \\
5 - \frac{5}{10}(v - 30) & \text{if } 30 < v \le 40 \\
0 & \text{if } v > 40
\end{array} $ \\\\
TMJ-R $D_{\text{mean}}$ \text{(Gy)} &
$ \text{Score} = \Biggl\lbrace \renewcommand{\arraystretch}{0.7}\begin{array}{@{}l@{\quad}l@{}}
3 & \text{if } v \le 0 \\
3 - \frac{0.5}{30}v & \text{if } 0 < v \le 30 \\
2.5 - \frac{2.5}{10}(v - 30) & \text{if } 30 < v \le 40 \\
0 & \text{if } v > 40
\end{array} $ \\\\
TMJ-L $D_{\text{mean}}$ \text{(Gy)} &
$ \text{Score} = \Biggl\lbrace \renewcommand{\arraystretch}{0.7}\begin{array}{@{}l@{\quad}l@{}}
3 & \text{if } v \le 0 \\
3 - \frac{0.5}{30}v & \text{if } 0 < v \le 30 \\
2.5 - \frac{2.5}{10}(v - 30) & \text{if } 30 < v \le 40 \\
0 & \text{if } v > 40
\end{array} $ \\\\
Parotid-R $D_{\text{mean}}$ \text{(Gy)} &
$ \text{Score} = \Biggl\lbrace \renewcommand{\arraystretch}{0.7}\begin{array}{@{}l@{\quad}l@{}}
3 & \text{if } v \le 0 \\
3 - \frac{0.5}{21}v & \text{if } 0 < v \le 21 \\
2.5 - \frac{2.5}{4}(v - 21) & \text{if } 21 < v \le 25 \\
0 & \text{if } v > 25
\end{array} $ \\\\
Parotid-L $D_{\text{mean}}$ \text{(Gy)} &
$ \text{Score} = \Biggl\lbrace \renewcommand{\arraystretch}{0.7}\begin{array}{@{}l@{\quad}l@{}}
3 & \text{if } v \le 0 \\
3 - \frac{0.5}{21}v & \text{if } 0 < v \le 21 \\
2.5 - \frac{2.5}{4}(v - 21) & \text{if } 21 < v \le 25 \\
0 & \text{if } v > 25
\end{array} $ \\\\
Lens-R $D_{1\%}$ \text{(Gy)} &
$ \text{Score} = \Biggl\lbrace \renewcommand{\arraystretch}{0.7}\begin{array}{@{}l@{\quad}l@{}}
3 & \text{if } v \le 0 \\
3 - \frac{0.5}{6}v & \text{if } 0 < v \le 6 \\
2.5 - \frac{2.5}{4}(v - 6) & \text{if } 6 < v \le 10 \\
0 & \text{if } v > 10
\end{array} $ \\\\
Lens-L $D_{1\%}$ \text{(Gy)} &
$ \text{Score} = \Biggl\lbrace \renewcommand{\arraystretch}{0.7}\begin{array}{@{}l@{\quad}l@{}}
3 & \text{if } v \le 0 \\
3 - \frac{0.5}{6}v & \text{if } 0 < v \le 6 \\
2.5 - \frac{2.5}{4}(v - 6) & \text{if } 6 < v \le 10 \\
0 & \text{if } v > 10
\end{array} $ \\\\
Eye-R $D_{\text{mean}}$ \text{(Gy)} &
$ \text{Score} = \Biggl\lbrace \renewcommand{\arraystretch}{0.7}\begin{array}{@{}l@{\quad}l@{}}
1.8 & \text{if } v \le 0 \\
1.8 - \frac{0.3}{30}v & \text{if } 0 < v \le 30 \\
1.5 - \frac{1.5}{10}(v - 30) & \text{if } 30 < v \le 40 \\
0 & \text{if } v > 40
\end{array} $ \\\\
Eye-L $D_{\text{mean}}$ \text{(Gy)} &
$ \text{Score} = \Biggl\lbrace \renewcommand{\arraystretch}{0.7}\begin{array}{@{}l@{\quad}l@{}}
1.8 & \text{if } v \le 0 \\
1.8 - \frac{0.3}{30}v & \text{if } 0 < v \le 30 \\
1.5 - \frac{1.5}{10}(v - 30) & \text{if } 30 < v \le 40 \\
0 & \text{if } v > 40
\end{array} $ \\\\
InnerEar-R $D_{\text{mean}}$ \text{(Gy)} &
$ \text{Score} = \Biggl\lbrace \renewcommand{\arraystretch}{0.7}\begin{array}{@{}l@{\quad}l@{}}
2.4 & \text{if } v \le 0 \\
2.4 - \frac{0.4}{30}v & \text{if } 0 < v \le 30 \\
2 - \frac{2}{10}(v - 30) & \text{if } 30 < v \le 40 \\
0 & \text{if } v > 40
\end{array} $ \\\\
InnerEar-L $D_{\text{mean}}$ \text{(Gy)} &
$ \text{Score} = \Biggl\lbrace \renewcommand{\arraystretch}{0.7}\begin{array}{@{}l@{\quad}l@{}}
2.4 & \text{if } v \le 0 \\
2.4 - \frac{0.4}{30}v & \text{if } 0 < v \le 30 \\
2 - \frac{2}{10}(v - 30) & \text{if } 30 < v \le 40 \\
0 & \text{if } v > 40
\end{array} $ \\

\end{longtable}
} 

\end{document}